\documentclass[runningheads]{llncs}
\usepackage[T1]{fontenc}
\usepackage{graphicx}
\usepackage{booktabs}
\usepackage{url}
\usepackage{epsfig}
\usepackage{graphicx}
\usepackage{amsmath}
\usepackage{amssymb}
\usepackage{booktabs}
\usepackage{algorithm}
\usepackage{algpseudocode}
\usepackage{dsfont}
\usepackage{amsmath,amssymb}
\usepackage{wrapfig}

\usepackage{tikz}
\usepackage{comment}
\usepackage{amsmath,amssymb} 
\usepackage{color}
\usepackage{tabularx}
\usepackage{multirow}
\usepackage{booktabs}
\usepackage{makecell}
\usepackage{enumitem}
\setitemize{noitemsep,topsep=0pt,parsep=0pt,partopsep=0pt}
\usepackage{caption}

\usepackage{color}

\usepackage{enumitem}
\usepackage[breaklinks,colorlinks,citecolor=green]{hyperref}
\usepackage{bbding}

\begin{document}
\title{MambaReg: Mamba-Based Disentangled Convolutional Sparse Coding for Unsupervised Deformable Multi-Modal Image Registration}
%
\titlerunning{MambaReg}
%

\author{Kaiang Wen \and
Bin Xie \and
Bin Duan \and
Yan Yan} 

\institute{
Department of Computer Science, Illinois Institute of Technology, Chicago, USA
\texttt{\{kwen2, bxie9, bduan2\}@hawk.iit.edu,\quad yyan34@iit.edu}\\
}

\authorrunning{Wen et al.}
%
%
\maketitle              

\begin{abstract}

Precise alignment of multi-modal images with inherent feature discrepancies poses a pivotal challenge in deformable image registration.
Traditional learning-based approaches often consider registration networks as black boxes without interpretability. 
One core insight is that disentangling alignment features and non-alignment features across modalities bring benefits. 
Meanwhile, it is challenging for the prominent methods for image registration tasks, such as convolutional neural networks, to capture long-range dependencies by their local receptive fields. 
The methods often fail when the given image pair has a large misalignment due to the lack of effectively learning long-range dependencies and correspondence. 
%
In this paper, we propose MambaReg, a novel Mamba-based architecture that integrates Mamba's strong capability in capturing long sequences to address these challenges. 
With our proposed several sub-modules, MambaReg can effectively disentangle modality-independent features responsible for registration from modality-dependent, non-aligning features. 
By selectively attending to the relevant features, our network adeptly captures the correlation between multi-modal images, enabling focused deformation field prediction and precise image alignment. 
The Mamba-based architecture seamlessly integrates the local feature extraction power of convolutional layers with the long-range dependency modeling capabilities of Mamba. 
Experiments on public non-rigid RGB-IR image datasets demonstrate the superiority of our method, outperforming existing approaches in terms of registration accuracy and deformation field smoothness.

\end{abstract}
\section{Introduction}
Image registration is a fundamental task in the computer vision field. Multi-modal image registration~(MIR) aims to spatially align two images from different modalities, which capture the same scene from different perspectives, such as pairs~\cite{yan2013cross} of RGB images that provide intensity and texture details and infrared images that give information on thermal properties that are not visible in the RGB spectrum. This alignment is crucial for subsequent multi-modal tasks like image fusion~\cite{hou2020vif} and segmentation~\cite{wang2018depth}. However, these multi-modal images are acquired by different sensors, they are often misaligned and contain both features responsible and not responsible for alignment, requiring further attention. 

Classical registration methods mostly involve an iterative optimization of similarity metrics like mutual information~\cite{wells1996multi}, correlation ratio~\cite{roche1998correlation}, and structural feature descriptors~\cite{bodensteiner2010local,heinrich2012mind} to maximize the similarity between modalities. These classical methods suffer from time-consuming and poor performance compared with deep-learning-based models. Recently, with the success of deep learning, many learning-based MIR methods have emerged and achieved remarkable success in inference speed~\cite{balakrishnan2019voxelmorph} and registration accuracy, broadly classified into two categories. One category~\cite{simonovsky2016deep,de2019deep,kang2022dual} trains deep networks to optimize certain similarity metrics for multi-modal registration. For example, de Vos et al.~\cite{de2019deep} proposed a twin CNN to minimize the normalized cross-correlation metric between multi-modal images for registration. However, these methods heavily rely on the effectiveness of similarity metrics. The other category~\cite{qin2019unsupervised,mahapatra2018deformable,arar2020unsupervised} adopts image-to-image translation to reduce MIR to a simpler mono-modal registration task. Arar et al.~\cite{arar2020unsupervised} employed a GAN to translate one modality to another, and then performed registration within the same modality. However, training GANs is challenging and may introduce artificial noise\cite{jabbar2021survey}, impeding registration.

These methods consider the networks as black-boxes and fail to provide better interpretability and relationship between modalities. One core insight is disentangling features that are responsible for alignment and features that are not responsible for alignment from different modalities is beneficial for multi-modal registration. This raises an intriguing question: how to build registration models capable of recognizing different features for alignment and non-alignment? 

Meanwhile, the prominent methods for image registration tasks, such as convolutional neuron networks, do not have the ability to capture long-range dependencies by their local receptive fields. The methods often fail when the given image pair has a large misalignment due to CNNs being lack of effectively learn long-range dependencies and correspondences. Existing methods typically require that image pairs are roughly aligned. Although vision transformers can exploit long-range dependencies by their attention mechanism, they suffer from high quadratic complexity and the requirements of massive data. Therefore, another question is raised: how to equip registration models with the ability to exploit long-term dependencies without too much resource consumption?

To address the aforementioned issues, we propose a Mamba-based multi-modal image registration network, named MambaReg, for deformable multi-modal registration tasks. Our proposed MambaReg consists of two interacted branches to disentangle alignment and non-alignment features from different modalities by our proposed modules. In order to exploit long-term dependencies, we integrate Mamba's~\cite{gu2023mamba} long sequence modeling capabilities into our model.  
In summary, our contributions in this paper are as follows:
\begin{itemize}[leftmargin=*]
    \item We propose a Mamba-based multi-modal image registration network, named MambaReg, for deformable multi-modal registration tasks. To the best of our knowledge, the proposed MambaReg is the first unsupervised Mamba-based image registration framework.

    \item We proposed different modules to disentangle modality-independent features responsible for registration from modality-dependent, non-aligning features.

    \item To address data scarcity in this domain, we reconstructed a plant RGB-IR registration dataset using the publicly available MSU-PID dataset.

    \item Extensive experiments on non-grid RGB-IR image datasets demonstrate that MambaReg outperforms existing methods and achieves state-of-the-art performance. 
    
\end{itemize}

\section{Related Works}

\noindent \textbf{Multi-modal Image Registration}
Multi-modal image registration (MIR) aims to align images from different modalities, which is more challenging than mono-modal registration due to the complex correspondence across different modalities. Several deep-learning-based approaches have been explored for rigid MIR. Nguyen et al.~\cite{nguyen2018unsupervised} developed a fast deep convolutional neural network for aligning multi-modal images with illumination variations by minimizing photometric loss. Pielawski et al.~\cite{pielawski2020comir} introduce contrastive learning to MIR that transfers MIR to a mono-modal task. 
Shao et al.~\cite{shao2021localtrans} employed a multi-scale transformer network to determine correspondences between cross-resolution images and achieve accurate homography estimation with up to a 10× resolution gap. Traditional deformable MIR methods optimize similarity metrics like mutual information (MI)~\cite{viola1997alignment}, and modality-independent neighborhood descriptor (MIND)~\cite{heinrich2012mind}. However, the process of optimizing the similarity metrics is usually iterative and time-consuming. To tackle this challenge, Wu et al.~\cite{wu2013unsupervised} proposed the first deep network to learn an application-specific similarity metric for multi-modal registration. Wang et al.~\cite{wang2019fire} bypassed domain translation by learning an encoder-decoder module to generate modality-independent features, which were then fed to a spatial transformer network(STN)~\cite{jaderberg2015spatial} to learn deformable transformations. Recently, generative adversarial network (GAN) based image-to-image translation methods have shown state-of-the-art performance in deformable MIR. Qin et al.~\cite{qin2019unsupervised}'s unsupervised multi-modal image registration network (UMDIR) based on disentangled image representation. Arar et al.~\cite{arar2020unsupervised} proposed a method combining an STN and a translation network for RGB/depth and RGB/NIR multi-modal registration, trained with a novel scheme alternating two different flows for spatial transformation. However, training GANs is challenging, and image translation may introduce artificial noises detrimental to registration~\cite{xu2020adversarial}. Li et al.~\cite{li2023multimodal} embed different modalities of images into the common feature space based on disentangled representations. Recently, Dent et al.~\cite{deng2023interpretable} proposed InMIR, which introduces the learned convolutional sparse coding(LCSC) to MIR for better interpretability.

\noindent \textbf{Mamba}
State Space Sequence Models (SSMs)~\cite{gu2023modeling} is a type of system that maps a 1-dimensional function or sequence $u(t) \rightarrow y(t) \in R$ and can be represented as the following linear Ordinary Differential Equation (ODE):
\begin{align}
&x'(t) =  Ax(t) +Bu(t) \\
&y(t) =  Cx(t)
\label{equa:ssm}
\end{align}
where state matrix $A \in \mathbb{R}^{N\times N}$ and $B,C \in \mathbb{R}^{N}$ are its parameters and $x(t) \in \mathbb{R}^{N}$ denotes the implicit latent state. SSMs offer several desired properties, such as linear computational complexity per time step and parallelized computation for efficient training. 
Despite their linear complexity per time-step and capacity for parallel computation, SSMs are generally memory-intensive and prone to vanishing gradients, which limits their use in sequence modeling.
The introduction of Structured State Space Sequence Models (S4)~\cite{gu2021efficiently} has mitigated these issues by imposing structured forms on the state matrix A and leveraging the High-Order Polynomial Projection Operator (HIPPO)~\cite{gu2020hippo} for initialization, thereby building deep sequence models with rich capability and efficient long-range reasoning ability. As a new type of network architecture, S4 has surpassed Transformers~\cite{vaswani2017attention} on the challenging Long Range Arena Benchmark~\cite{tay2020long}. Recently, Mamba~\cite{gu2023mamba} further advances SSMs in discrete data modeling such as text and genome. Mamba introduces an input-dependent selection mechanism, distinct from the traditional SSMs, and therefore achieves efficient information filtering from inputs. Besides, Mamba employs a hardware-aware algorithm, scaling linearly in sequence length, to compute the model recurrently with a scan. Mamba is faster than previous methods on modern hardware. In addition, The Mamba architecture, merging SSM blocks with linear layers, is notably simpler and has demonstrated state-of-the-art performance in various long-sequence domains.
\section{Proposed Method}
\begin{figure}[!t]
    \centering
    \includegraphics[width=\textwidth]{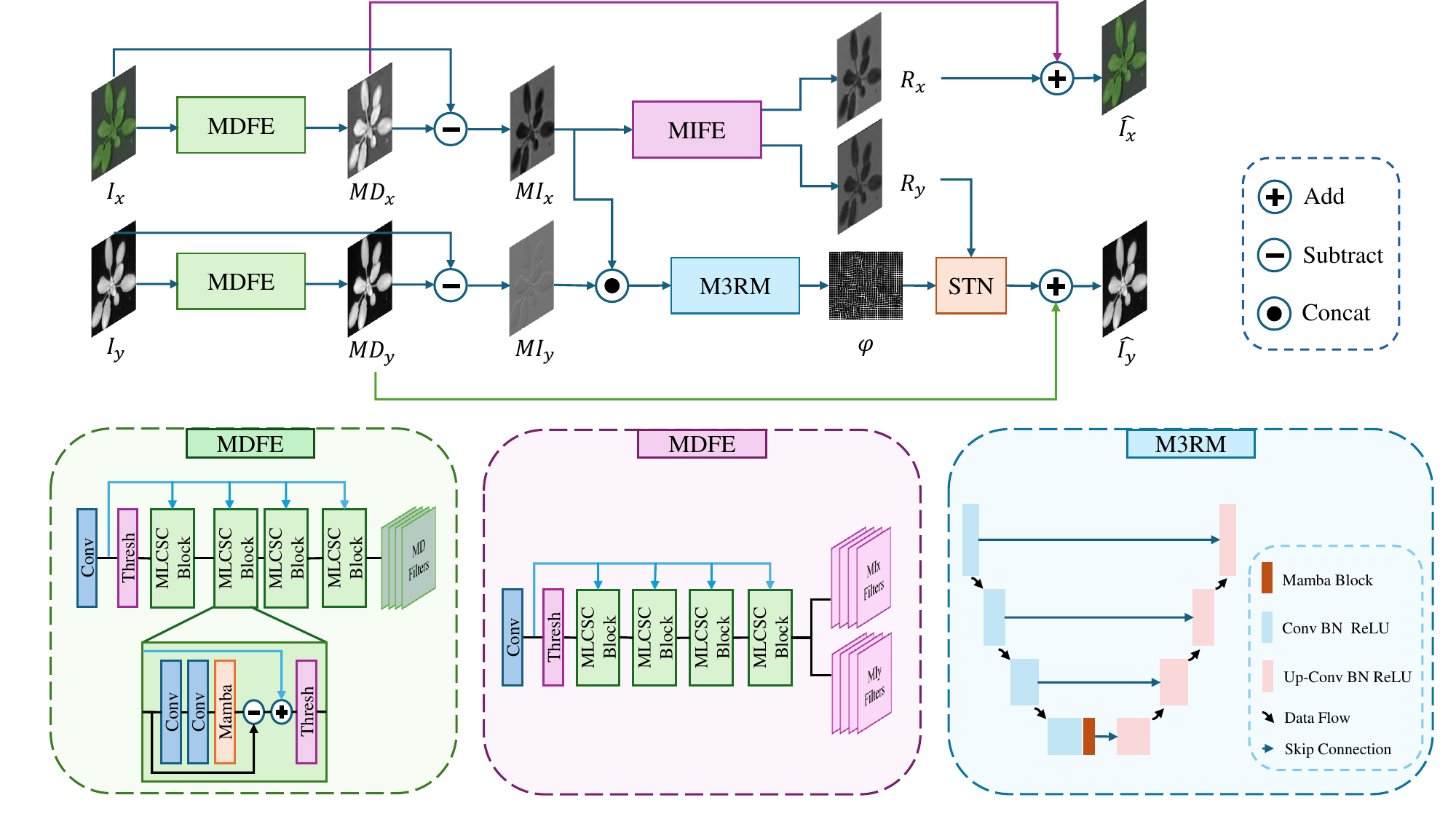}
    \caption{The framework of the proposed MambaReg for unsupervised multi-modal deformable image registration.} 
    \label{fig:framework}
\end{figure}

The network architecture of our proposed MambaReg is shown in Figure~\ref{fig:framework}. To perform multi-modality deformable registration, MambaReg consists of three modules, including the Modality-Dependent Feature Extractor(MDFE), Modality-Invariant Feature Extractor(MIFE) and Mamba-based Multi-Modal Registration Module(M3RM). In multi-modal image registration(MIR), not all features contribute positively to registration; some even hinder the effectiveness of registration due to the disparity in information content between different modalities. The key insight is to disentangle Modality-Invariant(MI) features responsible for the registration process from Modality-Dependent(MD) features and perform registration on MI features only. 
To disentangle MI and MD features, we utilize two MDFE modules to handle the moving image $I_x$ and the fixed image $I_y$ from two different modalities, respectively. The MDFE modules aim at extracting MD features from each modality $MD_x$ and $MD_y$ that do not contribute to the following image registration process. Therefore, MI features $MI_x$ and $MI_y$ could be obtained through the subtraction of ${MD_x,MD_y}$ from ${I_x,I_y}$. 
The proposed M3RM module takes the concatenated MI features ${MI_x, MI_y}$ as input to predict the registration field $\varphi$. The MIFE module aims to further encode the modality-invariant features from the moving image $MI_x$ and generate its reconstructed representations in two modalities $R_x, R_y$. To obtain the moved modality-independent features, $R_x$ is warped by the registration field $\varphi$ by applying a spatial transformer network(STN)~\cite{jaderberg2015spatial}. The MD feature of the fixed image $MD_y$ is then added to obtain the reconstructed fixed image $\hat{I_y}$. Segmentation labels of the moving and fixed image remain unseen during MambaReg’s training phase to ensure the network learns in an unsupervised manner.
Next, we will introduce the design of the three modules and training strategies.

\subsection{Modality-Dependent Feature Extractor(MDFE)} 
\label{sec:MDFE}

In multi-modal image registration, not all features contribute positively to registration; some even hinder the effectiveness of registration due to the disparity in information content between different modalities. Therefore, the MDFE module aims to extract the features that are dependent on the image modality $MD_x, MD_y$, which are not responsible for image registration and could be subtracted from the original image to obtain clean information. As depicted in Fig \ref{fig:framework}, each modality has a dedicated MDFE: one for the moving image modality $x$, and the other for the fixed image modality $y$. The moving image and the fixed image are denoted as $I_x$, and $I_y$ respectively. The MDFE module for moving image $I_x$ is designed to extract modality-dependent features of modality $x$ from the moving image. Initially, $I_x$ undergoes a convolutional operation to produce preliminary feature representations, which are then passed through a soft thresholding layer to enhance salient features. After this, a sequence of Mamba-based learned convolutional sparse coding (MLCSC) blocks is applied to generate Modality-Dependent Convolutional Sparse Representations (MDCSRs). Each MLCSC block comprises the Learned Convolutional Sparse Coding (LCSC) algorithm~\cite{sreter2018learned} with the bidirectional Mamba (Bi-Mamba) layer~\cite{zhu2024vision}, allowing for the capture of long-range dependencies within the image data. The selective SSM in the Bi-Mamba layer is beneficial to distinguish modality-dependent features from modality-independent features, therefore generating better MDCSRs. The MDCSRs are then convolved with Modality-Depent Filters(MD Filters), which are convolutional dictionary filters to capture modality-dependent features. The MDFE module for fixed image $I_y$ takes the fixed image as input and extracts modality-dependent features of modality $y$ in the same manner.



\subsection{Modality-Invariant Feature Extractor(MIFE)}
The Modality-Invariant Feature Extractor (MIFE) module aims to generate reconstructed modality-independent representations that capture the essential features shared across different modalities, facilitating effective multi-modal image registration. As depicted in Figure~\ref{fig:framework}, the MIFE architecture mirrors the MDFE module, comprising a series of MLCSC blocks and modality-specific filters. However, instead of operating on the original input images, the MIFE takes the modality-independent representations $MI_x$ as input, which is obtained by subtracting the moving image $I_x$ by the output of MDFE $MD_x$ . Through the sequence of MLCSC blocks, which incorporate the LCSC algorithm and bidirectional Mamba layers, the MIFE captures long-range dependencies and distinguishes modality-invariant features from modality-dependent ones. The selective SSM in the Bi-Mamba layers aids in this process, enabling the generation of Modality-Envariant Convolutional Sparse Representations (MICSRs). Subsequently, these MICSRs are convolved with modality-invariant filters for two modalities, namely $MI_x$ Filters and $MI_y$ Filters, to reconstruct the modality-independent representations $R_x$ and $R_y$ from the respective modalities.

The reconstructed representations $R_x$ and $R_y$, obtained through the MIFE module, contain the essential modality-invariant features shared across the input modalities, thereby facilitating effective multi-modal image registration by focusing on the common information content rather than modality-specific characteristics.

\subsection{Mamba-based Multi-Modal Registration Module(M3RM)}
The Mamba-based Multi-Modal Registration Module(M3RM) takes the MI features $MI_x$ and $MI_y$ from two modalities to predict the deformation field $\varphi$ for the multi-modal registration tasks, as shown in Fig~\ref{fig:framework}. The M3RM module comprise a U-Net~\cite{ronneberger2015u} architecture and a Mamba block at the last layer of the U-Net encoder. This design aims at integrating the local feature extraction power of convolutional layers with Mamba's impressive ability in long-term dependencies to extract features critical for registration. 

\subsection{Two-stage Training Strategy}
\label{sec:AGNet}
\begin{figure}[!t]
    \centering
    \includegraphics[width=0.95\textwidth]{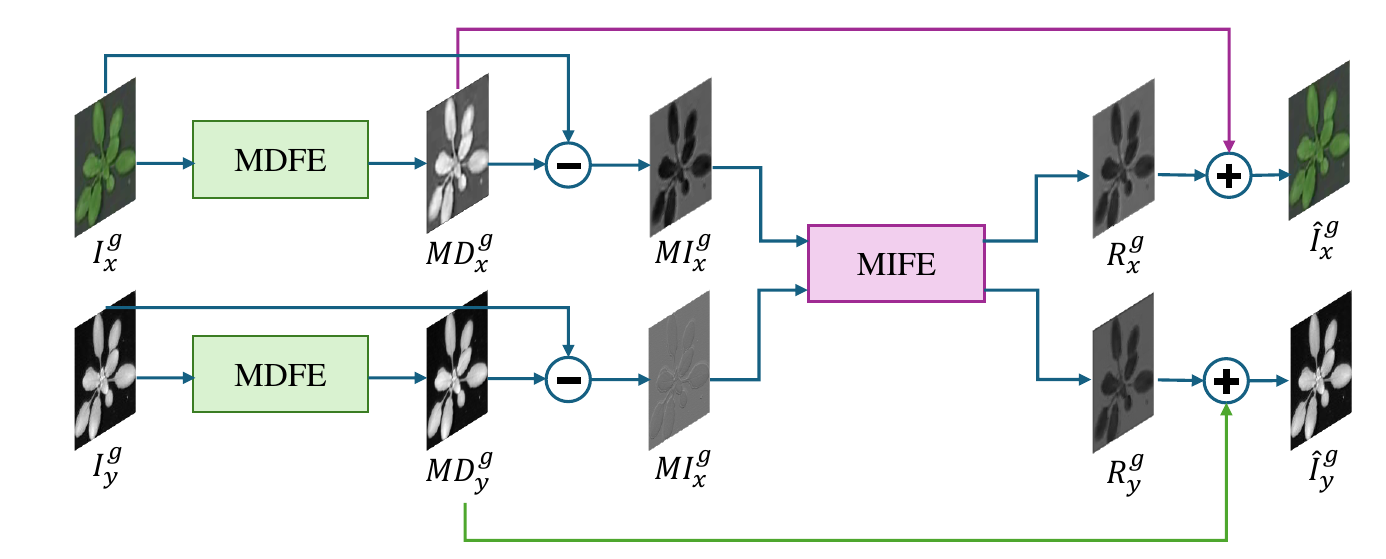}
    \caption{The architecture of accompanying guidance network (AG-Net) pre-trained on fully aligned multi-modal image pairs.} 
    \label{fig:AGNet}
\end{figure}

Following InMIR~\cite{deng2023interpretable}, we adopt a two-stage training strategy to better disentangle MI features from MD features through pre-training the AG-Net on fully registered multi-modal image pairs $I^g_x, I^g_y$ at the first stage.
As shown in Fig~\ref{fig:AGNet}, the AG-Net consists of 2 MDFE modules and 1 MIFE module. The MDFEs extract MD features $MD^g_x, MD^g_y$ from the two modalities. MI features are then computed by subtracting the corresponding MI features from the original input image:
\begin{align}
&MI^g_x = I^g_x - MD^g_x
&MI^g_y = I^g_y - MD^g_y
\label{equa:AGNet_MI}
\end{align}
Given MI features $MI^g_x, MI^g_y$ as input, the MIFE module then reconstructs modality-independent representations $R^g_x, R^g_y$ through a series of MLCSC blocks and modality-specific filters. Since the AG-Net is trained on fully aligned images, we obtain the reconstructed moving and fixed image $\hat{I}^g_x, \hat{I}^g_y$ by simply adding the MD features $MD^g_x, MD^g_y$ to reconstructs representations $R^g_x, R^g_y$.

The AG-Net is pre-trained by the following loss function:
\begin{align}
&\mathcal{L_A} = {\Vert \hat{I}^g_x - I^g_x \Vert}^2_2 + {\Vert \hat{I}^g_y - I^g_y \Vert}^2_2
\label{equa:AGNet_Loss}
\end{align}
Since $I^g_x, I^g_y$ are fully aligned, the reconstructs representations $R^g_x, R^g_y$ are obtained through the MIFE module without a registration network. During the second stage of training, the pre-trained AG-Net are used to guide the MI feature extraction in MambaReg through a guidance loss in the overall loss function, as described in Sec.~\ref{sec:Loss}.

\subsection{Binary ROI Mask}
\label{sec:ROI}
\begin{figure}[!t]
    \centering
    \includegraphics[width=0.9\textwidth]{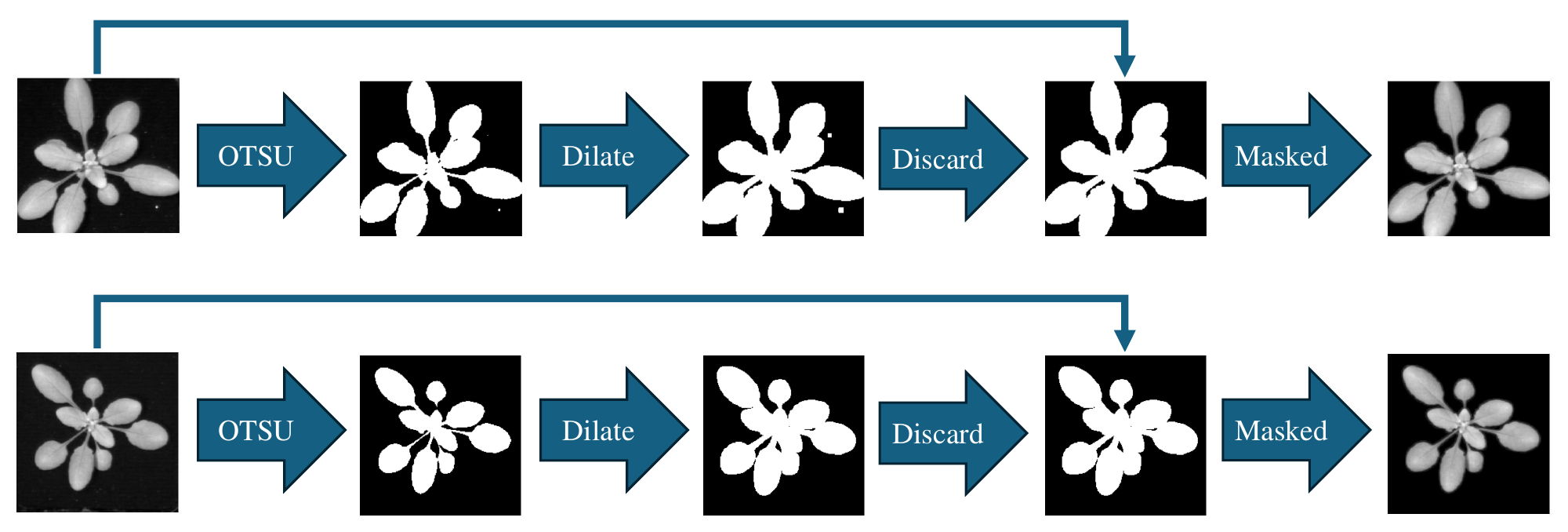}
    \caption{The generation process of the ROI mask for MSU-PID Dataset.} 
    \label{fig:ROI}
\end{figure}

To enable the model to focus more on the region of interest (ROI) during training and inference, we generate ROI masks for MSU-PID dataset~\cite{cruz2016multi} that highlight the object of interest while suppressing the background regions in an unsupervised manner. As shown in Fig~\ref{fig:ROI}, the ROI mask is obtained through a series of digital image processing techniques. This mask is generated based on the Otsu binarization method~\cite{otsu1975threshold}, which automatically determines an optimal threshold value for separating the object and background pixels in the image. The Otsu method assumes that the image contains two classes of pixels (foreground and background), and it calculates the optimal threshold value that minimizes the intra-class variance of the two classes. By applying this threshold, we obtain a binary mask where the pixels belonging to the object of interest are assigned a value of 1, and the background pixels are assigned a value of 0. After that, we apply dilation to the binary mask using a rectangular structural element to connect nearby regions and enlarges them, ensuring that significant but small features, like the branches and stems of the plant, are included in the foreground. Finally, we filtered out small, potentially irrelevant components using a preset pixel size. This step aims at discarding noise and minor artifacts. We will discuss how we incorporate the ROI mask in our end-to-end training process in the next section.

\subsection{Loss Function}
\label{sec:Loss}
The overall loss function for MambaReg, denoted as $\mathcal{L}$, is a weighted sum of four component losses: the ROI mask weighted similarity loss $\mathcal{L}_{sim}$, the smooth loss $\mathcal{L}_{smooth}$, the guidance loss $\mathcal{L}_G$, and the reconstruction loss $\mathcal{L}_R$. 

\textit{1) ROI Mask Weighted Similarity Loss: } To achieve better alignment in multi-modal deformable image registration, we utilize the mean squared error (MSE) to measure the similarity of the warped image $\hat{I}_y$ and the ground-truth image $I_gt$. As described in Sec.~\ref{sec:ROI}, plant could be a small object in the image during the early stages of plant growth. Therefore, the generated ROI masks are utilized as weights to the similarity loss to encourage the model to focus more on the region of interest instead of the background. We define the ROI mask weighted similarity loss as follows:
\begin{align}
\mathcal{L}_{sim} = \frac{1}{2}(  \mathrm{MSE}(x_{warp},\mathrm{GT}) + \mathrm{MSE}(\mathrm{mask}\times x_{warp},\mathrm{mask}\times \mathrm{GT})) 
\label{equa:L_sim}
\end{align}
where $x_{warp}$ is obtained by warping $I_x$ by the predicted registration field $\varphi$ through a STN~\cite{jaderberg2015spatial}.

\textit{2) Smooth Loss: } To encourage smooth deformation, we adopt the smooth loss in VoxelMorph~\cite{balakrishnan2019voxelmorph} to restrain large gradients on the predicted deformation field $\varphi$ as follows,
\begin{align}
\mathcal{L}_{smooth} = \sum_{\mathrm{p\in \varphi}  }  {\Vert \nabla \mathrm{p}  \Vert} ^2
\label{equa:L_smooth}
\end{align}
where $\mathrm{p}$ indicates the pixel in deformation field $\varphi$.

\textit{3) Guidance Loss: } As described in Sec.~\ref{sec:AGNet}, we utilize the MI features extracted from the pre-trained AG-Net as supervision to the MI feature extraction of the MambaReg-Net to achieve improvements in registration accuracy. The guidance loss is defined by utilizing the MSE loss to encourage similar MI features extracted from AG-Net and MambaReg-Net, following InMIR~\cite{deng2023interpretable}.

\textit{4) Reconstruction Loss: } As shown in Fig.~\ref{fig:framework}, MambaReg reconstructs the moving and fixed images as $\hat{I}_x, \hat{I}_y$. To achieve consistency in such reconstruction process, the reconstruction loss is defined as,
\begin{align}
\mathcal{L}_{R} = \mathrm{MSE}(\hat{I}_x,I_x) + \mathrm{MSE}(\hat{I}_y,I_y)
\label{equa:L_smooth}
\end{align}

\textit{5) Overall Loss: } The overall loss function for MambaReg $\mathcal{L}$ is a weighted sum of the above four component losses:
\begin{align}
\mathcal{L} = \alpha\mathcal{L}_{sim} + \beta\mathcal{L}_{smooth} + \gamma\mathcal{L}_G + \delta\mathcal{L}_{R}
\label{equa:L_smooth}
\end{align}
where $\alpha$, $\beta$, $\gamma$ and $\delta$ are the weights to each loss component.

\section{Experimental Results}

\subsection{Dataset}
\begin{figure}[!t]
    \centering
    \includegraphics[width=\textwidth]{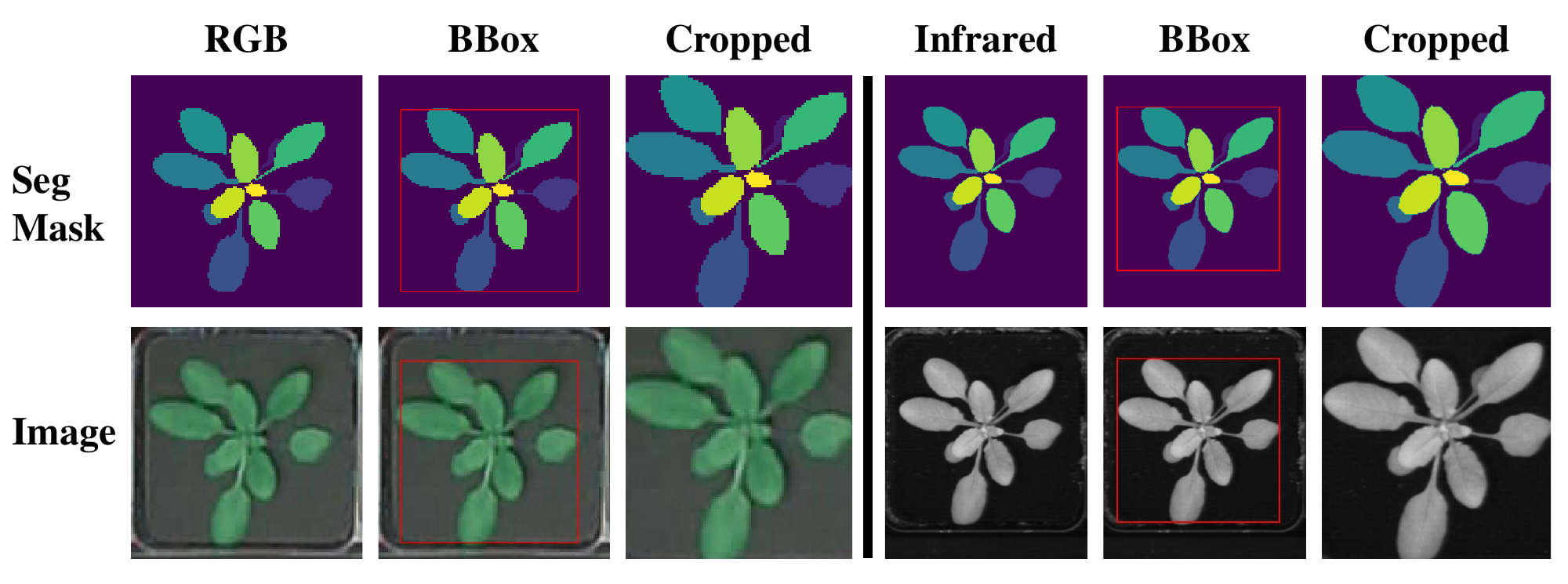}
    \caption{The reconstruction process of the MSU-PID dataset. After cropping, center alignment is achieved between RGB-IR image pairs, and the surrounding container is removed from the image.} 
    \label{fig:dataset}
\end{figure}
Based on the public multi-modality imagery database for plant phenotyping(MSU-PID) collected by Cruz et al.~\cite{cruz2016multi}, we have re-produced a non-rigid RGB-IR plant imagery dataset for deformable multi-modal image registration tasks. 
The original MSU-PID dataset comprises imagery of 16 Arabidopsis plants captured over 9 days, from 9 am to 11 pm, with each day yielding 15 frames. Out of these frames, 4 per day are given manually annotated instance segmentation labels. Each image contains number of instances varying from 4 to 14,  reflecting the natural growth progression of the plants.
However, directly utilizing the MSU-PID dataset for multi-modal image registration presents several challenges: (1) the RGB-IR pairs from the same plant and timestamp are not pre-aligned, (2) the growth of the plant results in new leaves, leading to variations between the instances in moving and fixed image pairs, and (3) there is significant background interference from containers.

\begin{figure}[t]
    \centering
    \includegraphics[width=0.95\textwidth]{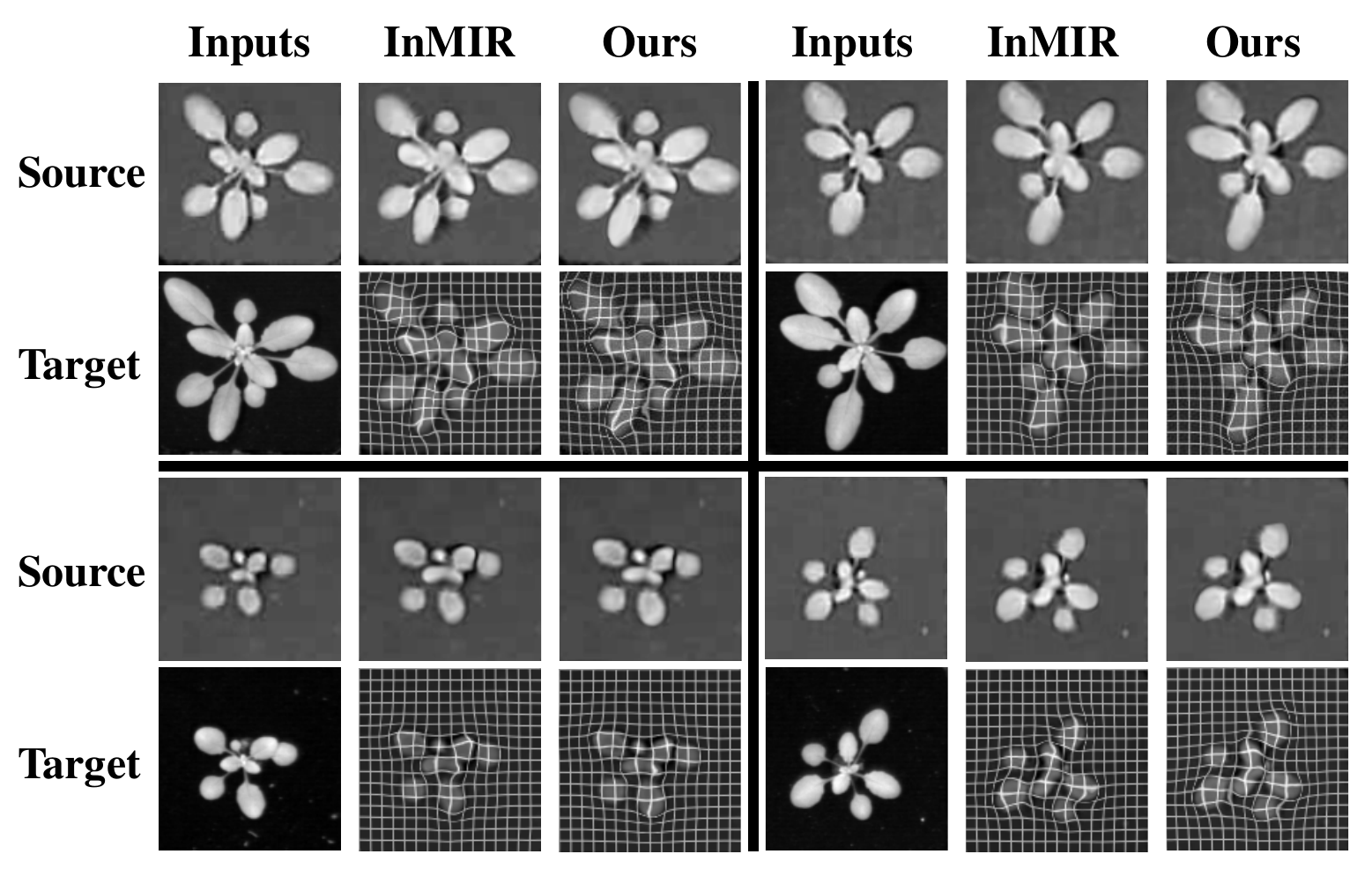}
    \caption{The deformable registration results on MSU-PID dataset.} 
    \label{fig:compare}
\end{figure}

To address these issues, we reconstruct the dataset as follows. First, we use the segmentation labels of the temporally latest frame of each plant to generate plant-specific bounding boxes. Using these bounding boxes, we cropped the RGB and infrared images, along with their corresponding labels. This allows aligning the center of RGB-IR image pairs and focusing solely on the plant regions instead of the surrounding containers. Then, from the cropped data, we selected RGB-IR image pairs of the same plant at different timestamps, allowing large deformation while ensuring that both modalities contained the same number of instances. 12,275 image pairs without annotations are available for the train set in unsupervised image registration, while 1,633 image pairs with annotations are available for the test set and computation of the Dice score~\cite{dice1945measures}. Finally, we randomly selected 300 out of the 12,275 pairs for training, and 900 out of the 1,633 pairs for testing. Besides, as mentioned in Sec.~\ref{sec:ROI}, we also generate the ROI binary mask for all RGB and infrared images in our reconstructed dataset. Our reconstructed dataset will be made publicly available.

\subsection{Implementation Details}
The network is trained using images with size 128 \texttimes 128. The number of MLCSC blocks in the MDFE and MIFE modules is set to 4. In the loss function, the $\alpha$, $\beta$, $\gamma$, and $\delta$ are set to 100, 10, 25, and 10, respectively. The total number of training epochs is 1,000. We use the Adam optimizer to train the network with a basic learning rate of 1e-4. PolyLR scheduler is used to adjust the learning rate during training based on the polynomial decay policy. We conduct our experiments on a single NVIDIA RTX A6000 GPU. The model is initialized with random weights with the random seed set as 3407. All implementations are done using the PyTorch deep learning framework.

\subsection{Evaluation Metrics}
For deformable registration, we adopt the Dice score~\cite{dice1945measures}, the mean square error(MSE), the normalized correlation coefficient(NCC) and SSIM to measure registration accuracy. The larger value of Dice score, NCC, and SSIM, and the smaller value of MSE indicate better performance in image registration. Since the leaf size varies in our reconstructed MSU-PID dataset, the computation of the Dice score is weighted according to the number of pixels of each label in the ground truth. The computation formula is as follows,
\begin{align}
\mathrm{Dice_{weighted}}  = \textstyle \sum_{i=1}^{N} \frac{\vert F_i \vert }{\sum_{i=1}^{N} \vert F_i \vert} \times \frac{ \vert F_i\cap W_i\vert}{\vert F_i\vert + \vert  W_i\vert} 
\label{equa:L_smooth}
\end{align}
where $N$ is the total number of labels, $F_i$ and $W_i$ are the pixel set of the $i$-th label in the segmentation of the fixed image and the warped image. The weighted Dice score provides a more balanced measurement of the registration performance.

\subsection{Quantitative and Qualitative Results}
\begin{table}[t!]
\begin{center}
\caption{The deformable RGB-IR image registration results on MSU-PID dataset in terms of MSE, DICE, NCC, and SSIM. The best results are marked in \textbf{bold}. 
}\label{tab:sota}
\vspace{2mm}
\begin{tabular}{l|c|c|c|c}
\toprule
Methods &  Dice$\uparrow$ &	MSE$\downarrow$ &	NCC$\uparrow$ &	SSIM$\uparrow$ \\

\midrule
Baseline & $72.09$ & $10.98\times10^{-3}$ & $76.75$ & $76.90$ \\
InMIR~\cite{deng2023interpretable} & $81.47$ & $57.52\times10^{-4}$ & $89.62$ & $83.34$ \\
MambaReg & $\mathbf{83.44}$ & $\mathbf{51.00\times10^{-4}}$ & $\mathbf{91.01}$ & $\mathbf{83.88}$ \\
\bottomrule
\end{tabular}
\end{center}
\end{table}

We compare our proposed approach with the state-of-the-art method, InMIR~\cite{deng2023interpretable}. Fig ~\ref{fig:compare} visualize the moved image and the corresponding deformation fields predicted by the proposed MambaReg and other baseline models. The qualitative comparison shows that MambaReg excels in registering image pairs with large deformations. The quantitative comparison with state-of-the-art results is illustrated in Table~\ref{tab:sota}. For the metrics of Dice, NCC, and SSIM, higher means better. For the metrics of MSE, lower means better. Table~\ref{tab:sota} illustrates that the number of Dice, NCC, and SSIM for our model is the highest, and the number of MSE is the lowest, which demonstrates the performance of our model is the best. We can conclude that our proposed method achieves the best performance in all metrics, which validates the effectiveness of our proposed model.


\subsection{Ablation Study}
\label{sec:ablation}

\begin{table}[t!]
\begin{center}
\caption{Ablation study of MDFE, MIFE, M3RM, Mamba type and the ROI mask. The best results are marked in \textbf{bold} and second bests are \underline{underlined}. }
\label{tab:ablation}
\vspace{2mm}
\begin{tabular}{c|c|c|c|c|c|c|c|c|c}
\toprule
& MDFE& MIFE& M3RM& Mamba Type & ROI Mask&  Dice$\uparrow$ &	MSE$\downarrow$ &	NCC$\uparrow$ &	SSIM$\uparrow$ \\

\midrule
B1 &\XSolidBrush & \XSolidBrush & \XSolidBrush & 1d Bi-Mamba & \XSolidBrush & \textbf{$78.78$} & $67.05\times10^{-4}$ & $87.32$ & $81.82$ \\
B2 &\CheckmarkBold & \XSolidBrush & \XSolidBrush & 1d Bi-Mamba & \XSolidBrush & \textbf{$80.12$} & $64.58\times10^{-4}$ & $87.86$ & $82.12$ \\
B3 &\CheckmarkBold & \CheckmarkBold & \XSolidBrush & 1d Bi-Mamba & \XSolidBrush & \textbf{$81.12$} & $58.22\times10^{-4}$ & $89.36$ & $83.07$ \\
B4 &\CheckmarkBold & \CheckmarkBold & \CheckmarkBold & 1d Mamba & \XSolidBrush & \textbf{$81.25$} & $56.96\times10^{-4}$ & $89.65$ & $83.29$ \\
B5 &\CheckmarkBold & \CheckmarkBold & \CheckmarkBold & 1d Bi-Mamba & \XSolidBrush & \textbf{$\underline{82.73}$} & $\mathbf{50.20\times10^{-4}}$ & $\mathbf{91.20}$ & $\mathbf{84.24}$ \\ \midrule
B6 &\CheckmarkBold & \CheckmarkBold & \CheckmarkBold & 1d Bi-Mamba & \CheckmarkBold & \textbf{$\mathbf{83.44}$} & $\underline{51.00\times10^{-4}}$ & $\underline{91.01}$ & $\underline{83.88}$\\

\bottomrule

\end{tabular}
\end{center}
\end{table}

\noindent\textbf{Baseline Models.} The proposed MambaReg has 6 baselines (\textit{i.e.}, B1, B2, B3, B4, B5, B6) as shown in Table \ref{tab:ablation}. All baselines contain three modules, MDFE, MIFE, and M3RM. The main differences are with or without different Mamba models, such as 1d Mamba or 1d Bi(-directional)-Mamba. (i) B1 are built without any Mamba-based layers. (ii) B2 are built without Mamba-based layers, except MDFE involving 1d Bi-Mamba. (iii) In B3, MDFE and MIFE involves 1d Bi-Mamba, but M3RM do not contain any Mamba-base layers. (iv) In B4, all modules, MDFE, MIFE and M3RM, involves 1d Mamba. (v)  In B5, all modules, MDFE, MIFE and M3RM, involves 1d Bi-Mamba. (vi) B6 is our full model, named MambaReg. All modules, MDFE, MIFE and M3RM, involves 1d Bi-Mamba. Meanwhile, B6 introduces binary ROI masks to focus more on the region of interest. 

\noindent\textbf{Ablation analysis.} The results of the ablation study are shown in~Table \ref{tab:ablation}. When we involve 1d Bi-Mamba into MDFE, MIFE and M3RM, the model improves  by 1.4\%, 2.4\% and 4.0\% in the DSC compared to B1. The result confirms the effectiveness that we involve 1d Bi-Mamba layers into MDFE, MIFE and M3RM modules. Compared to B4 that involves 1d Mamba into the three modules, B5 improves by 1.5\% in the DSC. The results confirms the effectiveness of 1d Bi-Mamba, which is better than 1d Mamba. When we introduce binary ROI masks into our model to focus more on the region of interest, B6 improves by 0.7\% in the DSC compared to B5. The results confirms the effectiveness of the introduction of ROI masks. Therefore, the results demonstrate the effectiveness of our proposed MambaReg.
\section{Conclusion}

MambaReg, the proposed Mamba-based architecture for unsupervised deformable multi-modal image registration, has demonstrated its superiority by disentanglement modality-dependent and modality-independent features and exploiting long-range dependencies within images. The specialized modules, i.e., the Modality-Dependent Feature Extractor and Modality-Invariant Feature Extractor enable the focused extraction and processing of relevant features necessary for precise registration. Extensive experiments on the reconstructed MSU-PID dataset have shown that MambaReg achieves state-of-the-art performance in registration accuracy.

Our results underscore the advantages of incorporating Mamba into the network structure, particularly in managing the complexities inherent in multi-modal datasets where alignment and non-alignment features must be distinctly processed. The capability of Mamba to handle long sequence data efficiently, with minimal computational overhead, positions MambaReg as a scalable solution suitable for various practical applications in medical imaging and beyond. Future work will focus on further optimizing these disentanglement strategies and exploring the potential of Mamba-based models in other domains of computer vision where similar challenges exist.

{\small
\bibliographystyle{ieee_fullname}
\bibliography{egbib}
}

\end{document}